\documentclass[final,1p,times,authoryear]{elsarticle}
\usepackage[authoryear,round]{natbib}

\usepackage{amsmath}  
\usepackage{amssymb}  
\usepackage{amsfonts} 
\usepackage{algorithmic}
\usepackage{textcomp}
\usepackage{mathtools}
\usepackage{graphicx}    
\usepackage{stfloats}     
\usepackage{lipsum}       
\usepackage{multirow}
\usepackage{array}
\usepackage{booktabs}
\usepackage{lmodern} 
\usepackage{xltabular} 
\usepackage{xcolor}
\usepackage{color}
\usepackage{enumitem}
\usepackage{float}
\usepackage[colorlinks,bookmarksopen,bookmarksnumbered,citecolor=black, linkcolor=black, urlcolor=black]{hyperref}
\usepackage{colortbl} 

\journal{Computer Vision and Image Understanding}

\begin{document}

\begin{frontmatter}

\title{BTMTrack: Robust RGB-T Tracking via Dual-template Bridging and Temporal-Modal Candidate Elimination} 

\author[label1]{Zhongxuan Zhang}
\author[label1]{Bi Zeng\corref{cor1}}
\author[label1]{Xinyu Ni}
\author[label1]{Yimin Du}
\cortext[cor1]{Corresponding author. Email: zb9215@gdut.edu.cn}
\affiliation[label1]{organization={Guangdong University of Technology},
            addressline={No. 100 Waihuan Xi Road, Guangzhou Higher Education Mega Center}, 
            city={Guangzhou},
            postcode={510006}, 
            state={Guangdong},
            country={China}}

\begin{abstract}
RGB-T tracking leverages the complementary strengths of RGB and thermal infrared (TIR) modalities to address challenging scenarios such as low illumination and adverse weather. However, existing methods often fail to effectively integrate temporal information and perform efficient cross-modal interactions, which constrain  their adaptability to dynamic targets. In this paper, we propose BTMTrack, a novel framework for RGB-T tracking. The core of our approach lies in the dual-template backbone network and the Temporal-Modal Candidate Elimination (TMCE) strategy. The dual-template backbone effectively integrates temporal information, while the TMCE strategy focuses the model on target-relevant tokens by evaluating temporal and modal correlations, reducing computational overhead and avoiding irrelevant background noise. Building upon this foundation, we propose the Temporal Dual Template Bridging (TDTB) module, which facilitates precise cross-modal fusion through dynamically filtered tokens. This approach further strengthens the interaction between templates and the search region. Extensive experiments conducted on three benchmark datasets demonstrate the effectiveness of BTMTrack. Our method achieves state-of-the-art performance, with a 72.3\% precision rate on the LasHeR test set and competitive results on RGBT210 and RGBT234 datasets.
\end{abstract}




\end{frontmatter}

\section{Introduction}
\label{sec:intro}
Visual Object Tracking (VOT) is a key task in computer vision that aims to localize a target in consecutive video frames. While many approaches~\citep{gao2023generalized,chen2022backbone,zeng2024visual,ye2022joint} have achieved impressive performance under standard conditions, RGB-based trackers often struggle in complex scenarios such as low illumination, cluttered backgrounds, and adverse weather conditions. These challenges, stemming from the reliance of visible-light imaging on environmental conditions, significantly constrain their robustness in real-world applications. Thermal infrared (TIR) imaging, which captures heat emissions, serves as a complementary modality to RGB data, enabling reliable performance in low-light or obstructed scenarios. By leveraging the strengths of both RGB and TIR modalities, RGB-T tracking has shown great potential in diverse applications such as surveillance~\citep{alldieck2016context} and autonomous driving~\citep{dai2021tirnet}.
\begin{figure}[!htb]
\centering 
\includegraphics[width=0.7\linewidth]{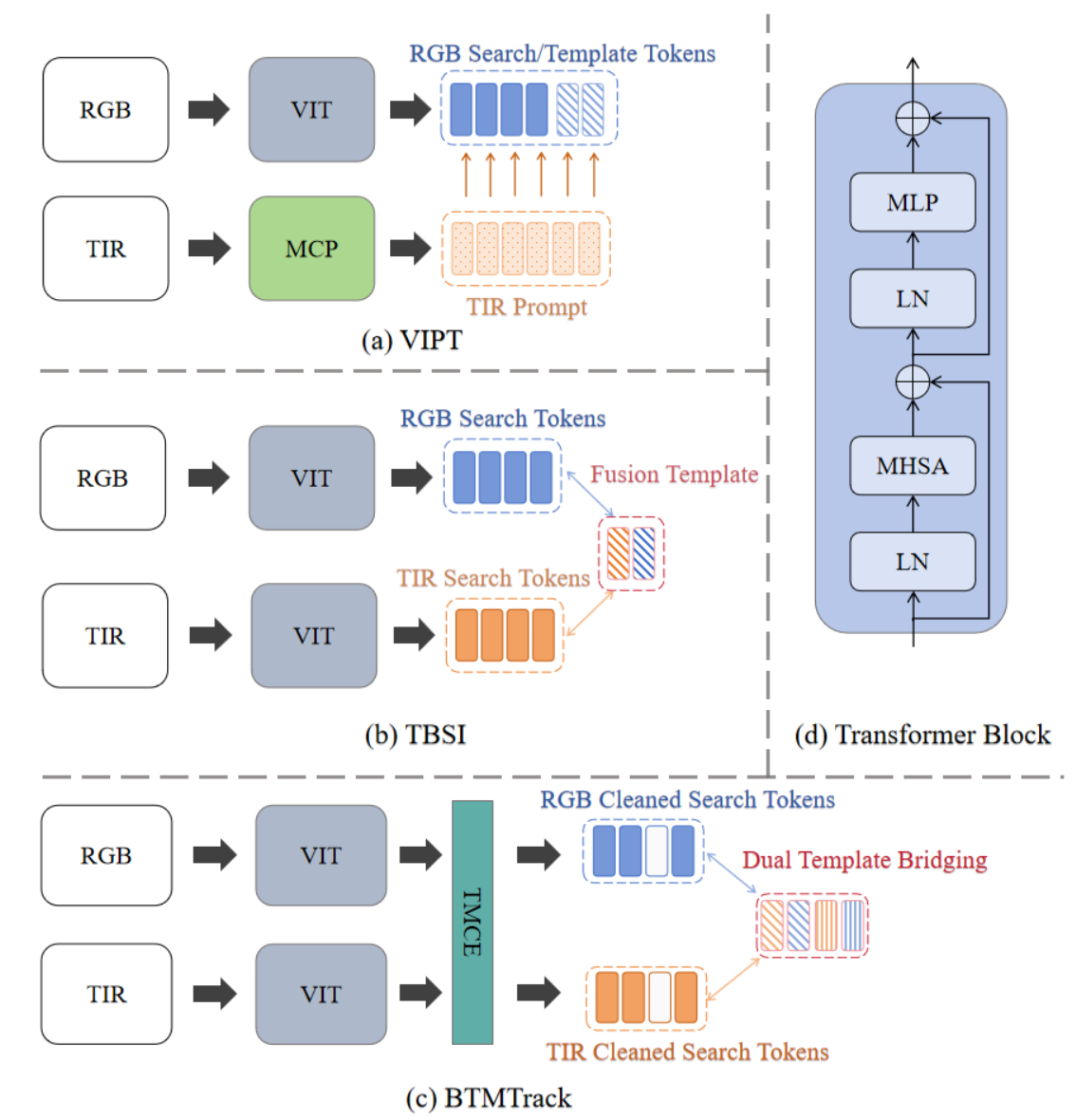}
\caption{Comparison of our cross-modal fusion approach with previous methods. (a) VIPT, injects TIR modality information as prompt-based auxiliary input into the RGB modality network. (b) TBSI, uses template tokens as a bridge to mediate interactions between the search regions of the two modalities. (c) Our model, filters target-relevant search region tokens before performing dual-temporal template bridging. (d) Example of a Transformer block.} 
\label{fig:fig1}
\end{figure}

The primary goal of RGB-T tracking is to effectively fuse the complementary information from RGB and TIR modalities, ensuring robust tracking in diverse environments. Recent advancements, including ViPT~\citep{zhu2023visual} and TBSI~\citep{hui2023bridging}, have demonstrated notable success by proposing innovative strategies for multi-modal feature fusion. As shown in Fig.\ref{fig:fig1} (a) and (b), ViPT~\citep{zhu2023visual} utilizes TIR as prompts for RGB to enhance feature interaction, while TBSI~\citep{hui2023bridging} employs a Template-Bridged Search region Interaction structure for cross-modal fusion. However, these approaches often overlook the importance of leveraging temporal information to address target variations, and they fail to effectively mitigate the impact of background noise, which reduces the model's discriminative capability. Therefore, two significant challenges still remain. The first challenge lies in handling target variations during tracking, such as changes in aspect ratio, deformation, or fast motion. Without considering temporal information, trackers face difficulties in adapting to these dynamic changes, resulting in issues such as tracking drift or inaccurate localization. The second challenge is the impact of background noise, which introduces redundancy and misguides the model’s judgment. Conventional fusion strategies often fail to effectively suppress irrelevant information, further weakening the tracker’s ability to accurately distinguish the target from the background.

To address the challenges of dynamic target variations, some recent works, such as TATrack~\citep{wang2024temporal}, have incorporated temporal information by introducing online template updates. These methods enable the tracker to adapt to target variations by leveraging historical information during tracking process. However, they primarily focus on limited template interactions, failing to fully exploring the synergy between static and dynamic templates. As a result, these approaches often fail to strike a balance between capturing target variations and maintaining stable representation, which can lead to tracking drift in complex scenarios. To tackle the aforementioned challenges, we propose BTMTrack, a novel RGB-T tracking framework that integrates dynamic templates as a core component of its design. Static and dynamic templates interact within a unified framework, where self-attention mechanisms across transformer blocks. This design enables the tracker to effectively balance object appearance and recent changes, thereby enhancing its adaptability to dynamic target variations. By balancing these two perspectives, our framework achieves robust performance in tracking scenarios involving significant target variations. While dynamic templates enhance the tracker's adaptability to target variations, they are insufficient on their own to effectively address the challenge of background noise. This prompts a critical question: can we design an approach that simultaneously leverages temporal information, models cross-modal associations, and mitigates background interference without incurring excessive computational overhead?

Inspired by recent advancements such as BAT~\citep{cao2024bi} and USTrack~\citep{xia2023unified}, we observe that the dominant modality should adapt dynamically to varying scenarios rather than relying exclusively on either RGB or TIR modalities. For instance, RGB images excel in well-lit environments with rich color and texture, while TIR provides higher confidence in low-visibility conditions. To address this, we propose the Temporal-Modal Candidate Elimination (TMCE) strategy, which dynamically prunes noisy tokens in the search region by leveraging both temporal and modal features. By jointly evaluating the contributions of RGB and TIR modalities, TMCE mitigates single-modality reliance and selectively retains the most effective tokens. Furthermore, it incorporates scores from both static and dynamic templates to handle target variations, leveraging temporal and multi-modal features for precise token selection. TMCE reduces the computational cost of dynamic templates while suppressing background noise, thereby enhancing robustness in complex environments. To complement TMCE, we introduce the Temporal Dual Template Bridging (TDTB) module, which builds upon and extends the template-bridged search region interaction inspired by TBSI~\citep{hui2023bridging}. Through bidirectional attention mechanisms, TDTB enables static and dynamic templates to collaboratively interact with the search region, deeply fusing RGB and TIR information while enriching templates with contextual relevance. Together, TMCE and TDTB collaboratively adapt to temporal and modal variations, ensuring robust and efficient performance in RGB-T tracking.
The primary contributions of this work are summarized as follows:

\begin{itemize}
  \item We propose BTMTrack, an innovative RGB-T tracking framework that leverages both static and dynamic templates to enable adaptive temporal modeling and robust object tracking.
  \item We develop the TMCE strategy, which conducts temporal and modality-aware token pruning, effectively suppressing background noise while ensuring computational efficiency.
  \item We propose the TDTB module, which facilitates comprehensive cross-modal interactions between templates and search regions, thereby improving feature fusion and enhancing overall robustness.
  \item Our tracker achieves state-of-the-art performance across three benchmarks, especially attaining a precision of 72.3\% on the LasHeR test set, while also delivering competitive results on RGBT210 and RGBT234.
\end{itemize}

\section{Related Works}
\subsection{RGB-T Tracking}
RGB-based trackers often struggle in challenging scenarios such as low illumination, thermal crossover, and adverse weather due to their reliance on visible light. By leveraging the complementary strengths of visible (RGB) and thermal infrared (TIR) modalities, RGB-T tracking achieves robust performance across diverse environments. mfDiMP~\citep{zhang2019multi} incorporates multimodal fusion at multiple stages to align and combine features from RGB and TIR modalities. APFNet~\citep{xiao2022attribute} adopts an attribute-based progressive fusion network, aggregating features under specific conditions such as occlusion or illumination variation. SiamCDA~\citep{zhang2021siamcda} reduces modality differences before fusion, enhancing the quality of the fused features. While effective, these methods often struggle with background noise and irrelevant regions during fusion, limiting the complementary benefits of RGB and TIR modalities in complex scenarios.

In contrast to traditional approaches, Transformer architectures have significantly advanced RGB-T tracking by leveraging self-attention mechanisms for cross-modal interaction. DRGCNet~\citep{mei2023differential} and MIRNet~\citep{hou2022mirnet} use cross-attention to transfer discriminative features across modalities while filtering redundant information through gating mechanisms. TBSI~\citep{hui2023bridging} introduces a stacked extraction-fusion module to improve interaction between RGB and TIR features, whereas ViPT~\citep{zhu2023visual} employs lightweight fusion modules to simplify the fusion process and reduce computational costs. However, these methods typically process all tokens in the search region indiscriminately, failing to prioritize target-relevant features, which introduces noise and inefficiencies. In this paper, we propose the Temporal Dual Template Bridging (TDTB) module, which selectively processes tokens that are dynamically pruned through a temporal-modal evaluation process. Unlike existing methods, TDTB focuses on the most target-relevant tokens, jointly determined by RGB and TIR modalities. By unifying static and dynamic templates with these filtered tokens, TDTB facilitates precise and efficient cross-modal interaction, enabling the model to focus on critical features from both modalities. This design enhances the model's ability to capture complementary information between modalities while dynamically adapting to temporal variations, ensuring robust and efficient RGB-T tracking across diverse scenarios.

\subsection{Temporal Information Exploitation}

Temporal information is critical for robust visual tracking, enabling trackers to adapt to target variations over time. Existing works can be categorized into three main approaches: template update methods, temporal propagation strategies, and multi-modal temporal modeling. Template update methods aim to maintain dynamic templates that adapt to target appearance changes. For example, UpdateNet~\citep{zhang2019learning} uses additional networks to optimize template updates, STARK~\citep{yan2021learning} introduces a dual-template design with scoring-based dynamic updates, and MixFormer~\citep{cui2022mixformer} selects the highest-scoring frame as its dynamic template. However, these methods often treat temporal updates as isolated processes, neglecting the interplay between temporal and spatial information. Temporal propagation strategies emphasize utilizing historical information to guide future predictions. ARTrack~\citep{wei2023autoregressive} employs autoregressive frameworks to propagate motion dynamics but suffers from high complexity and computational overhead. TCTrack~\citep{cao2022tctrack} leverages online temporally adaptive convolution to refine spatial features with historical data but remains limited to single-modality tracking, missing cross-modal complementarities.

Multi-modal temporal modeling integrates temporal and cross-modal information. TATrack~\citep{wang2024temporal} introduces an online template for dual-branch tracking, leveraging temporal and modal complementarity. However, it primarily uses TIR to enhance RGB tracking without fully involving TIR in the tracking process. Additionally, its complex spatio-temporal interaction modules introduce significant computational costs. In this paper, we propose the Temporal-Modal Candidate Elimination (TMCE) strategy. TMCE dynamically prunes noisy tokens in the search region by synchronously evaluating temporal, cross-modal, and spatial features. By filtering background noise, it enhances target focus while leveraging temporal and cross-modal correlations. This approach ensures robust and efficient RGB-T tracking across diverse scenarios.

\begin{figure*}[!htb]
\centering 
\includegraphics[width=1\textwidth]{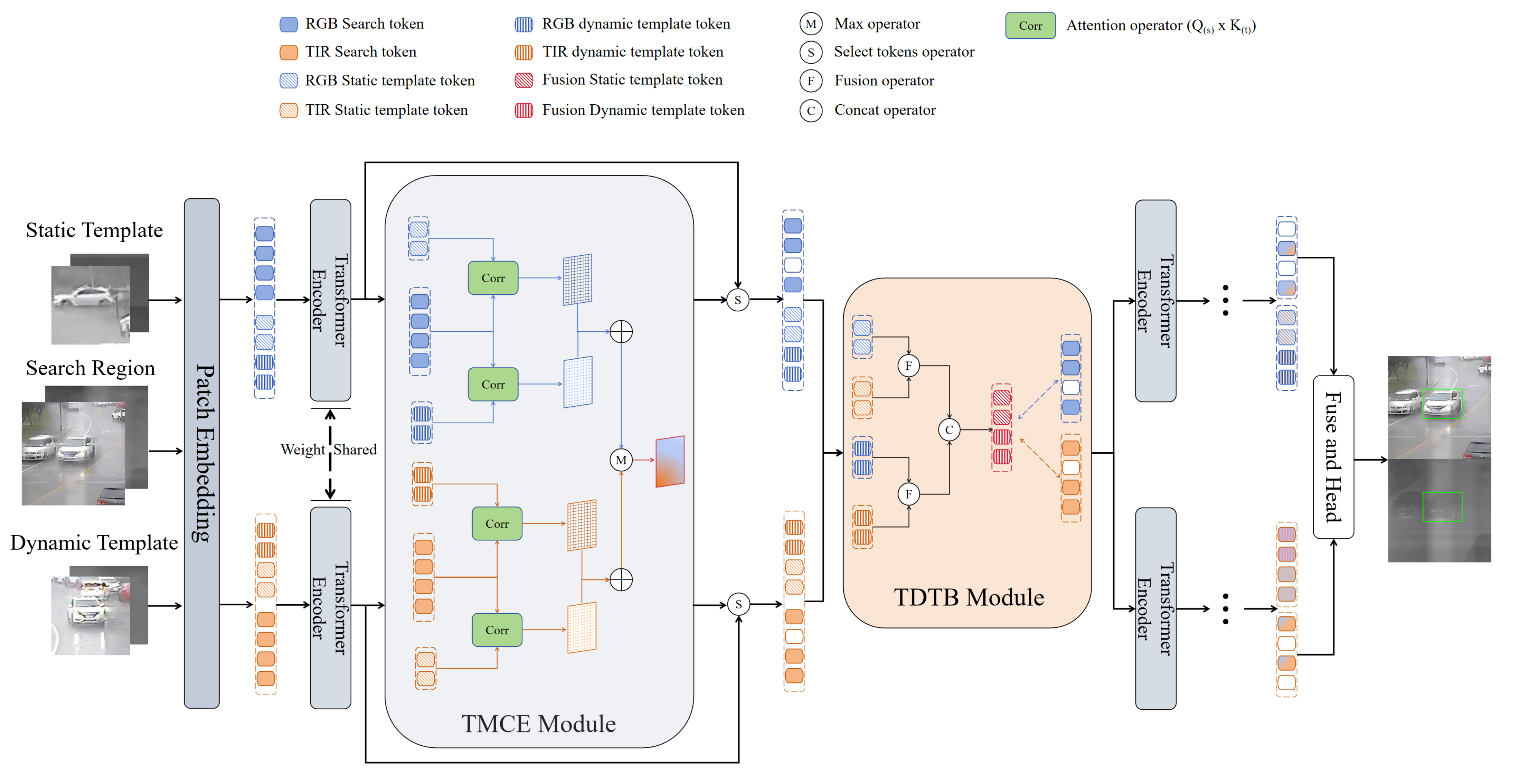}

\caption{The overall framework of our method. It integrates static and dynamic templates with search regions from RGB and TIR patches. These patches are tokenized and processed by a ViT backbone for feature extraction. The proposed TMCE strategy filters tokens based on temporal and modal relevance to reduce background noise. The TDTB module enables interactions between dual-temporal templates and search regions of both modalities. Finally, fused RGB and TIR features are passed to the tracking head to predict the target's location.}
\label{fig:fig2}
\end{figure*}

\section{Method}
\label{sec:method}
In this section, we introduce BTMTrack, an RGB-T tracking method that fully leverages temporal and modal information to address background noise challenges. The overall framework of BTMTrack is illustrated in Fig.\ref{fig:fig2}
\subsection{RGB-T Tracking Process Based on ViT}
Similar to other state-of-the-art RGB-T tracking methods like TBSI~\citep{hui2023bridging} and BAT~\citep{cao2024bi}, we follow the ViT-based framework introduced in VOT~\citep{ye2022joint,cui2022mixformer,lin2022swintrack} and use the target states and consecutive video frames from both modalities as inputs. Before feeding the data into the model, we also preprocess the dynamic templates specifically based on the current tracking state and input them into the framework accordingly.

Before the tracking process begins, the RGB-T tracker first extracts the appropriate templates $z_{\text{static}}^{\text{rgb}}, z_{\text{static}}^{\text{tir}} \in \mathbb{R}^{H_z \times W_z \times 3}$ from the target state of the first frame and simultaneously initializes the dynamic template $z_{\text{dynamic}}^{\text{rgb}}, z_{\text{dynamic}}^{\text{tir}}$, which is identical to the static template. These templates and their corresponding search regions, denoted as $x^{\text{rgb}}, x^{\text{tir}} \in \mathbb{R}^{H_x \times W_x \times 3}$, are then passed through the same Patch Embedding layer for tokenization as:
\begin{equation}
\begin{array}{c}
    x^{\text{rgb}}, x^{\text{tir}} \in \mathbb{R}^{N_x \times (3P^2)} \\
    z_{\text{static}}^{\text{rgb}}, z_{\text{static}}^{\text{tir}}, z_{\text{dynamic}}^{\text{rgb}}, z_{\text{dynamic}}^{\text{tir}} \in \mathbb{R}^{N_z \times (3P^2)}
\end{array}
\end{equation}

Here, \( P \times P \) represents the resolution of each patch, and \( N_z \) and \( N_x \) denote the number of patches in the template and search images, respectively.
Subsequently, they are enhanced with positional encodings, and the tokens from both the RGB and TIR modalities are then concatenated as:
\begin{equation}
I^{\text{rgb}} = \Bigl( z_{\text{static}}^{\text{rgb}}, z_{\text{dynamic}}^{\text{rgb}}, x^{\text{rgb}} \Bigr), \\
I^{\text{tir}} = \Bigl( z_{\text{static}}^{\text{tir}}, z_{\text{dynamic}}^{\text{tir}}, x^{\text{tir}} \Bigr)
\end{equation}
These are then input into a series of transformer blocks, where multi-modal joint feature extraction and selective feature fusion take place. 

After feature extraction through the backbone network, the token information from both modalities is converted into feature maps and undergoes a simple fusion through convolutional layers. The fused features are then fed into the prediction head to obtain the final output bounding box. The process is as follows:
\begin{equation}
B = H\left( \text{Conv}\left( F\left( I_{\text{rgb}}, I_{\text{tir}} \right) \right) \right)
\end{equation}
Where B represents the final predicted bounding box, H denotes the prediction head, Conv refers to the 3×3 convolutional layer, and F represents our ViT backbone network.

\subsection{Temporal-Modal Candidate Elimination}
\label{sec:tmce}
Transformer-based object tracking models possess strong global modeling capabilities, capturing long-range dependencies between the target and various regions in the scene, enabling accurate target localization even in complex scenarios. However, retaining all tokens is not necessary. Since the contributions of different tokens to target localization are uneven, a large number of tokens originating from the background or irrelevant regions often lack useful information, introduce noise and redundancy, and ultimately degrade model performance while increasing computational cost. Consequently, selecting key tokens and filtering irrelevant information is an effective strategy for improving both model efficiency and accuracy. Addressing the problem of selective token interaction, previous work~\citep{ye2022joint} proposed the CE mechanism, which leverages the correlation map from the attention computation to score and filter search region tokens, retaining only the highest-scoring tokens for interaction with template tokens. However, the CE method is limited to single-modality information and does not account for temporal dependencies.

To address this issue, we propose TMCE (as illustrated in Fig.\ref{fig:fig2}), which dynamically filters effective target candidate tokens based on temporal information and the synchronized evaluation of RGB and TIR modality features, thereby enhancing tracking accuracy and efficiency.

\textbf{Balanced Contribution of Dual Templates.} The static template provides global and stable features of the target, helping to anchor its original appearance, while the dynamic template captures temporal variations and adapts to changes in complex scenarios. Therefore, we believe that the static and dynamic templates contribute equally to feature recognition, and we directly add the two correlation maps, which are individually generated by the static template and the dynamic template with the search region, respectively. The process is formally defined as follows:
\[
Corr\_Map_{\text{static}} = \text{Softmax}\left(Q_{\text{static template}} K_{\text{search}}^\top\right)
\]
\[
Corr\_Map_{\text{dynamic}} = \text{Softmax}\left(Q_{\text{dynamic template}} K_{\text{search}}^\top\right) 
\]
\begin{equation}
Corr\_Map = \left(Corr\_Map_{\text{static}} + Corr\_Map_{\text{dynamic}}\right) 
\label{eq:eq4}
\end{equation}

\textbf{Spatial-level Adaptive Modality Selection.} Inspired by outstanding RGB-T tracking works such as BAT~\citep{cao2024bi} and USTrack~\citep{xia2023unified}, we observe that the confidence levels of RGB and TIR modalities vary significantly across different environments. Specifically, in well-lit and clear weather conditions, the RGB modality typically provides more detailed appearance features, such as color, texture, and target boundaries. In contrast, under nighttime, foggy, or high-glare conditions, the confidence of the RGB modality decreases, while the TIR modality becomes more reliable by capturing thermal radiation characteristics of the target.

Unlike other works that emphasize dominant-subordinate modality selection or modality weight allocation, we adopt a spatial perspective to select the modality with the higher score at each specific patch. We believe that this approach retains the most valuable spatial information from the perspectives of both modalities, enabling more accurate classification and localization tasks. By comparing the corresponding positions on the correlation maps of RGB and TIR ($Corr\_Map_{\text{ RGB}}, Corr\_Map_{\text{ TIR}}$), which are respectively derived through Eq.\ref{eq:eq4}, we obtain a new correlation map. The process is formally defined as follows:

\begin{equation}
{
Corr\_Map_{\text{ Final}} = \text{max}\left(Corr\_Map_{\text{ RGB}} + Corr\_Map_{\text{ TIR}}\right)
}
\end{equation}\\
Subsequently, we sort all search tokens by their scores on $Corr\_Map_{\text{ Final}}$, retaining the top k tokens with the highest scores. Our TMCE strategy effectively integrates temporal, spatial, and cross-modality information, fully exploiting the strengths of the dual-template baseline. It enhances the model's robustness while significantly reducing computational complexity.

\subsection{Temporal Dual Template Bridging Module}
\begin{figure*}[!htb]
\centering 
\includegraphics[width=1\textwidth]{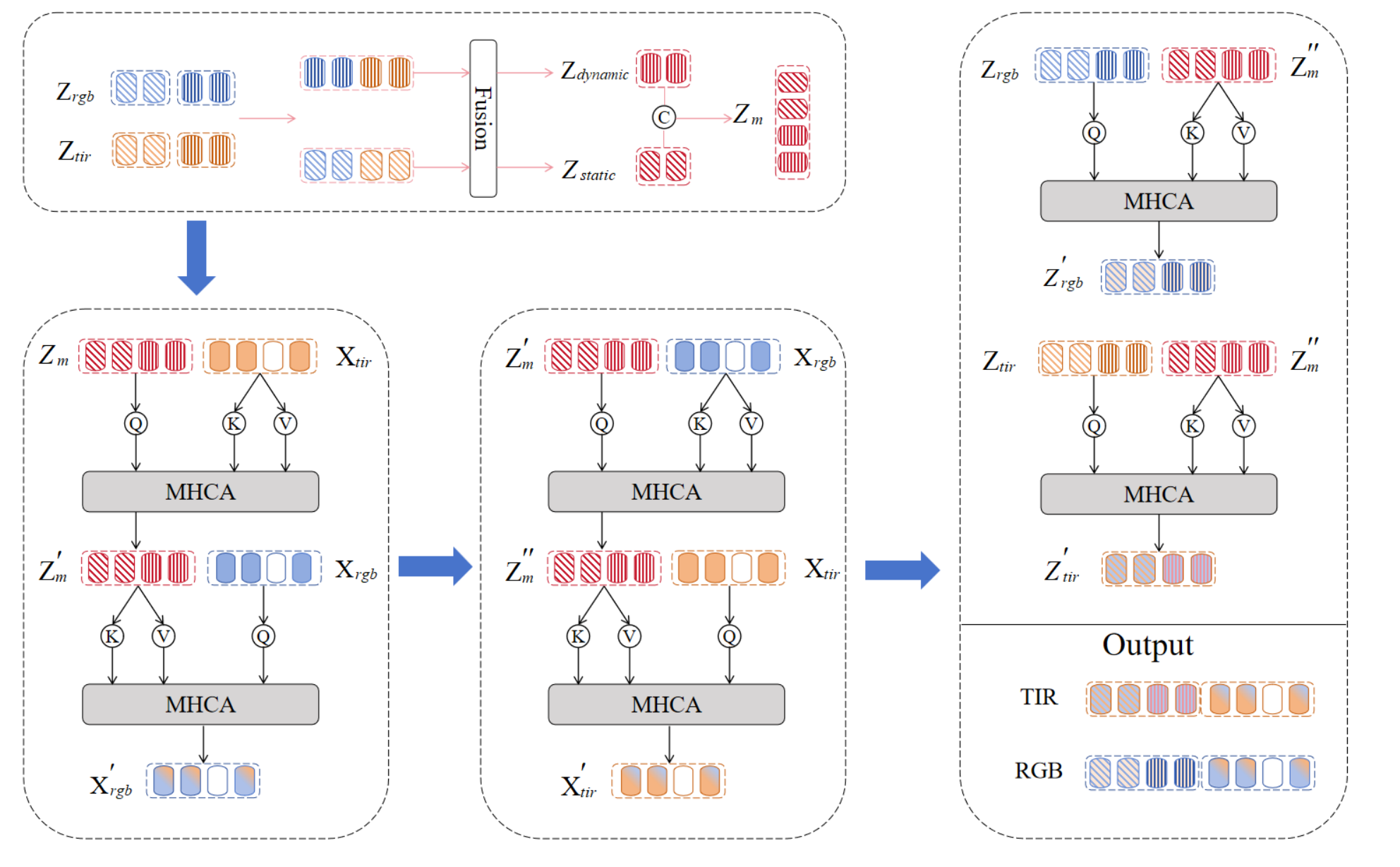}
\caption{The diagram demonstrates the process of dual-temporal template fusion and six MHCA operations within the TDTB module. For clearer presentation, we omit details such as LN, MLP, and the residual connections typically performed in each Transformer block.} 
\label{fig:fig3}
\end{figure*}

To complement the effects of TMCE, we propose the Temporal Dual Template Bridging (TDTB) module, which further optimizes the interaction between the static and dynamic templates. This module is inserted after the first feature selection to ensure effective information fusion on the basis of retaining the most representative features of both modalities. Inspired by the TBSI~\citep{hui2023bridging}, we adopt a similar template bridging search region interaction method for modality fusion, while effectively connecting the static and dynamic templates along the temporal dimension. By fully utilizing temporal information, TDTB helps the model better adapt to target variations and enhances its performance in dynamic scenarios. The data flow of this module is shown in Fig.\ref{fig:fig3}.

\textbf{Temporal Template Fusion.} To facilitate dynamic multimodal interaction across different temporal dimensions, we separately fuse the static and dynamic templates of the RGB and TIR modalities, resulting in two multimodal bridging intermediates: $Z_{static}\in \mathbb{R}^{N_z \times C}$ and $Z_{dynamic}\in \mathbb{R}^{N_z \times C}$, which capture temporal and modality-specific features. The formulas for this process are as follows:
\[
Z_{static} = [Z_{static}^{rgb}; Z_{static}^{tir}] W_m 
\]
\begin{equation}
Z_{dynamic} = [Z_{dynamic}^{rgb}; Z_{dynamic}^{tir}] W_m
\end{equation}\\
Where $W_m\in \mathbb{R}^{2C \times C}$ is the parameter of the linear layer used in the fusion operator. Subsequently, $Z_{static}$ and $Z_{dynamic}$ influence the fusion process between the two modalities from different temporal perspectives.

\textbf{Dual Template-Bridged Interaction with Search Regions.} To enable effective cross-modal interaction, we employ multi-head cross-attention between the template intermediates and their corresponding search regions. This process ensures that information from the static and dynamic templates is seamlessly integrated with the search regions of both RGB and TIR modalities. In the following equations, we will define multi-head cross-attention as $\text{MHCA}(Q, K, V)$, where Q, K, and V represent the query, key, and value matrices, respectively. The details of the Transformer block are shown in Fig.\ref{fig:fig1}(d). We define the MHCA operation as follows, omitting the multi-head operation for clarity:
\begin{equation}
\text{MHCA}(Q, K, V) = \text{Softmax}\left( \frac{({Q} {W}_q)({K} {W}_k)^T}{\sqrt{C}} \right)({V} {W}_v)
\end{equation}\\
Where \( W_q \), \( W_k \), and \( W_v \) are learnable parameter matrices that project the input features into the query, key, and value spaces, respectively. In the cross-attention mechanism of this module, K is equal to V, so the equation can be simplified as follows:
\begin{equation}
\text{MHCA}(X, Y) = \text{Softmax}\left( \frac{({X} {W}_q)({Y} {W}_k)^T}{\sqrt{C}} \right)({Y} {W}_v)
\end{equation}

First, we concatenate the $Z_{static}$ and $Z_{dynamic}$ directly to obtain $Z_{m} = [Z_{static}; Z_{dynamic}]$, which is then used in the subsequent interactions. Then, we update the template \( Z_{m} \) to a new template \( Z_{m}^{'} \) using the target-relevant contextual information from the TIR search region $X_{tir}$. The process is as follows:
\[
Z_{temp}^{'} = \text{LN}(Z_{m} + \text{MHCA}(Z_{m}, X_{tir}))
\]
\begin{equation}
Z_{m}^{'} = \text{LN}(Z_{temp}^{'} + \text{MLP}(Z_{temp}^{'}))
\end{equation}

After obtaining the target-relevant contextual information from the TIR search region $X_{tir}$, the updated temporal dual-template $Z_{m}^{'}$ will propagate the information to the RGB search region $X_{rgb}$ via MHCA, LN, and MLP. The detailed process is as follows:
\[
X_{rgb\_temp}^{'} = \text{LN}(X_{rgb} + \text{MHCA}(X_{rgb}, Z_{m}^{'})
\]
\begin{equation}
X_{rgb}^{'} = \text{LN}(X_{rgb\_temp}^{'} + \text{MLP}(X_{rgb\_temp}^{'}))
\end{equation}

Similarly, we use the updated temporal dual-template $Z_{m}^{'}$ as the new dual-template input, and by reversing the interaction order between the search region modalities (from TIR $\rightarrow$ RGB to RGB $\rightarrow$ TIR), we perform the same operations to obtain a new fused template $Z_{m}^{''}$, which has interacted with both search regions, as well as the updated TIR search region tokens $X_{tir}^{'}$ that have interacted with the RGB search region's contextual information, as illustrated in the flow of Fig.\ref{fig:fig3}.

\textbf{Original Templates Interaction with Medium.} Finally, we leverage the fused template $Z_{m}^{''}$, enriched with cross-modal information, to update the original RGB and TIR templates($Z_{rgb} = [Z_{static}^{rgb}; Z_{dynamic}^{rgb}]$ and $Z_{tir} = [Z_{static}^{tir}; Z_{dynamic}^{tir}]$). The core of this process lies in utilizing the comprehensive information provided by Z to fully explore the potential correlations between the RGB and TIR modalities, thereby enhancing the adaptability and accuracy of the templates in object tracking. By effectively extracting and interacting the multimodal information within the fused template, we can dynamically adjust each modality’s template, further improving the model’s robustness in complex environments. The process for obtaining the updated RGB template $Z_{rgb}^{'}$ is as follows:
\[
Z_{rgb\_temp}^{'} = \text{LN}(Z_{rgb} + \text{MHCA}(Z_{rgb}, Z_{m}^{''})
\]
\begin{equation}
Z_{rgb}^{'} = \text{LN}(Z_{rgb\_temp}^{'} + \text{MLP}(Z_{rgb\_temp}^{'}))
\end{equation}\\
Similarly, we can obtain the updated TIR template $Z_{tir}^{'}$, which will then be combined with $X_{tir}^{'}$, $Z_{rgb}^{'}$, and $X_{rgb}^{'}$ to serve as the output of the TDTB module.

\subsection{Prediction Head and Loss Function}
In this work, we utilize a unified prediction head design, consistent with prior methods~\citep{ye2022joint}. The prediction head transforms token sequences into a 2D spatial feature map via fully convolutional networks (FCN). This process outputs three components: the target classification score map (indicating the target location), the offset, and the normalized bounding box dimensions.

The total loss function is defined as:

\begin{equation}
L_{\text{total}} = L_{\text{cls}} + \lambda_1 L_{\text{iou}} + \lambda_2 L_1
\end{equation}

where \( L_{\text{cls}} \) represents the classification loss computed with weighted Focal Loss~\citep{ross2017focal}, and \( L_{\text{iou}} \) and \( L_1 \) are the bounding box regression losses, optimized using Generalized IoU Loss~\citep{rezatofighi2019generalized} and L1 Loss, respectively. The trade-off parameters \( \lambda_1 \) and \( \lambda_2 \) balance the contribution of these components.

\section{Experiments}
\subsection{Implementation Details}
Our model is implemented using PyTorch~\citep{paszke2019pytorch}, with all experiments conducted on two RTX 3090 GPUs. Additionally, speed tests were performed on a single RTX 2080Ti GPU.

\textbf{Architectures}. For the backbone network, we utilize the pre-trained Vision Transformer (ViT) model, which has been trained on the SOT task, as adopted in several advanced approaches~\citep{hui2023bridging,zhu2023visual,cao2024bi} in the RGB-T tracking domain. The TMCE strategy is applied after the 4th, 7th, and 10th Transformer blocks, while the TDTB module is inserted only after the first application of the TMCE strategy. The resolution of the template is set to $128\times128$ pixels, while the search region has a resolution of $256\times256$ pixels.

\textbf{Training}. Our model is trained on the LasHeR~\citep{li2021lasher} dataset, and evaluated on the LasHeR~\citep{li2021lasher}, RGBT210~\citep{li2017weighted}, and RGBT234~\citep{li2019rgb} datasets. We set the training to 15 epochs, with 60,000 image pairs per epoch. Each GPU handles 32 image pairs, resulting in a total batch size of 64. The learning rate for the ViT backbone is set to 1e-5, while the learning rate for the other parameters is set to 1e-4. After 10 epochs, the learning rate is decayed by a factor of 10.

\textbf{Inference}. During inference, the initial static template, dynamic template, and search region are used as inputs to the model. We determine whether to update the dynamic template based on the target classification score predicted by the prediction head.

\renewcommand{\arraystretch}{1.2}
\begin{table*}[!t]
\centering
\resizebox{\textwidth}{!}{
\begin{tabular}{c c c c c c c c c c c}
\toprule
\multirow{2}{*}{\raisebox{-0.5ex}{\fontsize{10pt}{12pt}\selectfont Method}} & \multirow{2}{*}{\raisebox{-0.5ex}{\fontsize{10pt}{12pt}\selectfont Source}} & \multirow{2}{*}{\raisebox{-0.5ex}{\fontsize{10pt}{12pt}\selectfont Backbone}} & \multicolumn{3}{c}{LasHeR} & \multicolumn{2}{c}{RGBT234} & \multicolumn{2}{c}{RGBT210}\\
\cmidrule(lr){4-6}\cmidrule(lr){7-8}\cmidrule(lr){9-10}
 & & & SR & PR & NPR & MSR & MPR & MSR & MPR \\
\midrule
MANet++~\citep{lu2021rgbt}   & TIP 2021   & VGG-M      & 31.7 & 46.7 & 40.8 & 55.4 & 80.0 & 55.3 & 78.5 \\
DMCNet~\citep{lu2022duality}    & TNNLS 2022 & VGG-M      & 35.5 & 49.0 & 43.1 & 59.3 & 83.9 & 55.9 & 79.7 \\
APFNet~\citep{xiao2022attribute}    & AAAI 2022  & VGG-M      & 36.2 & 50.0 & -    & 57.9 & 82.7 & 57.1 & 82.1 \\
CAT++~\citep{liu2024rgbt}     & TIP 2024   & VGG-M      & 35.6 & 50.9 & 44.4 & 59.2 & 84.0 & 56.1 & 82.2 \\
HMFT~\citep{zhang2022visible}      & CVPR 2022  & ResNet-50  & -    & -    & -    & 56.8 & 78.8 & 53.5 & 78.6 \\
CMD~\citep{zhang2023efficient}       & CVPR 2023  & ResNet-18  & 46.4 & 59.0 & 54.6 & 58.4 & 82.4 & 59.3 & 83.4 \\
MFNet~\citep{zhang2022rgb}     & IVC 2022   & ResNet-50  & 46.7 & 59.7 & 55.4 & 60.1 & 84.4 & -    & -    \\
mfDiMP~\citep{zhang2019multi}    & CVPRW 2019 & ResNet-50  & 46.7 & 59.9 & -    & 59.1 & 84.2 & 59.3 & 84.9 \\
VIPT~\citep{zhu2023visual}      & CVPR 2023  & ViT-Base   & 52.5 & 65.1 & 61.7 & 61.7 & 83.5 & 60.3 & 82.1 \\
QueryTrack~\citep{fan2024querytrack}& TIP 2024   & ViT-Base   & 52.0 & 66.0 & -    & 60.0 & 84.1 & -    & -    \\
OneTracker~\citep{hong2024onetracker}& CVPR 2024  & ViT-Base   & 53.8 & 67.2 & -    & 64.2 & 85.7 & -    & -    \\
TBSI~\citep{hui2023bridging}      & CVPR 2023  & ViT-Base   & \textcolor{green}{56.5} & 69.2 & 66.5 & 63.7 & 87.1 & \textcolor{blue}{62.5} & \textcolor{blue}{85.3} \\
BAT~\citep{cao2024bi}       & AAAI 2024  & ViT-Base   & 56.3 & \textcolor{green}{70.2} & 66.4 & 64.1 & 86.8 & -    & -    \\
TATrack~\citep{wang2024temporal}   & AAAI 2024  & ViT-Base   & 56.1 & \textcolor{green}{70.2} & \textcolor{green}{66.7} & \textcolor{green}{64.4} & \textcolor{green}{87.2} & \textcolor{green}{61.8} & \textcolor{blue}{85.3} \\
GMMT~\citep{tang2024generative}      & AAAI 2024  & ViT-Base   & \textcolor{blue}{56.6} & \textcolor{blue}{70.7} & \textcolor{blue}{67.0} & \textcolor{blue}{64.7} & \textcolor{blue}{87.9} & -    & -    \\
BTMTrack  & Ours       & ViT-Base   & \textcolor{red}{58.0} & \textcolor{red}{72.3} & \textcolor{red}{68.5} & \textcolor{red}{65.4} & \textcolor{red}{88.3} & \textcolor{red}{63.7} & \textcolor{red}{87.5} \\
\bottomrule
\end{tabular}
}
\caption{Comparison of our model with other state-of-the-art methods on the LasHeR, RGBT234, and RGBT210 benchmarks.
The top three results are highlighted in \textcolor{red}{red}, \textcolor{blue}{blue}, and \textcolor{green}{green}, respectively.}
\label{tab:tab1}
\end{table*}

\subsection{Comparison with State-of-the-art Methods}
To evaluate the performance and robustness of our model, we compare it with recent state-of-the-art RGB-T tracking models on three different RGB-T tracking benchmarks.

\textbf{LasHeR}. LasHeR~\citep{li2021lasher} is a large-scale and diverse RGB-T tracking dataset, consisting of 1224 RGB and thermal infrared video sequences, with a total of over 730K frames. The dataset’s test set includes 245 challenging video sequences, which capture RGB and thermal infrared image pairs under various scenes and conditions. The goal is to evaluate the performance of RGB-T tracking models in complex environments. We evaluate our model using three metrics: Success Rate(SR), Precision Rate(PR), and Normalized Precision Rate(NPR), and compare it with 15 other advanced trackers on the LasHeR dataset. The RGB-T trackers included in the comparison are: TATrack~\citep{wang2024temporal}, BAT~\citep{cao2024bi}, TBSI~\citep{hui2023bridging}, GMMT~\citep{tang2024generative}, OneTracker~\citep{hong2024onetracker}, QueryTrack~\citep{fan2024querytrack}, VIPT~\citep{zhu2023visual}, mfDiMP~\citep{zhang2019multi}, MFNet~\citep{zhang2022rgb}, CMD~\citep{zhang2023efficient}, CAT++~\citep{liu2024rgbt}, APFNet~\citep{xiao2022attribute}, DMCNet~\citep{lu2022duality}, and MANet++~\citep{lu2021rgbt} . The results, as shown in Tab.\ref{tab:tab1}, indicate that our model outperforms previous state-of-the-art methods on this dataset. Specifically, BTMTrack surpasses GMMT~\citep{tang2024generative} by 1.4\% in SR and 1.6\% in PR. The effectiveness of our model on the large-scale RGB-T dataset LasHeR~\citep{li2021lasher} clearly demonstrates that combining spatiotemporal information while avoiding irrelevant background noise can significantly improve performance.

\textbf{RGBT210}. RGBT210~\citep{li2017weighted} is a classic RGB-T tracking benchmark consisting of 210 video sequences, with a total of approximately 210K frames. Each video pair in the dataset contains up to 8K frames, offering long-term tracking challenges for RGB-T trackers. As shown in Tab.\ref{tab:tab1}, BTMTrack outperforms TBSI~\citep{hui2023bridging} by 1.2\% and 2.2\% in terms of Maximum Success Rate (MSR) and Maximum Precision Rate (MPR), respectively.

\textbf{RGBT234}. RGBT234~\citep{li2019rgb} is an extension of the RGBT210~\citep{li2017weighted} dataset, featuring additional environmental challenges. It consists of 234 video sequences, totaling approximately 234K frames. The dataset includes 12 attributes, such as Low Illumination (LI), Occlusion, Deformation (DEF), and Movement, providing a more diverse benchmark for evaluating RGB-T tracking models. Ground truth labels are provided for both RGB and thermal infrared (TIR) modalities, enabling performance evaluations across multiple modalities. As shown in Tab.\ref{tab:tab1}, BTMTrack achieves the best performance on the RGBT234~\citep{li2019rgb} dataset compared to 15 other advanced RGB-T trackers. It surpasses GMMT~\citep{tang2024generative} by 0.7\% in MSR and by 0.4\% in MPR, further validating the effectiveness of our model.

\subsection{Ablation Studies}
\textbf{Component analysis}. In Tab.\ref{tab:tab2}, we conducted ablation experiments on the individual components of our model using the LasHeR~\citep{li2021lasher} dataset, all under the same parameter settings. Additionally, to validate the effectiveness of the TDTB module, we included the TBSI module as a component for comparison in this ablation study.

We denote \textbf{\#1} as the baseline, which consists of the ViT backbone pre-trained on the SOT task dataset without any additional components. \textbf{\#2} adds the TBSI~\citep{hui2023bridging} module to the baseline. \textbf{\#3} further introduces a dynamic template enriched with temporal information into the backbone network, which is then concatenated and fed into the TBSI~\citep{hui2023bridging} module. \textbf{\#4} replaces the TBSI~\citep{hui2023bridging} module in \textbf{\#3} with our TDTB module. The results from \textbf{\#1} to \textbf{\#4} demonstrate that the introduction of dual templates can improve the model's performance to some extent; however, it also introduces a larger resource overhead, which slows down the model. Moreover, the TBSI~\citep{hui2023bridging} module cannot fully leverage the temporal information within the dual templates. In contrast, by replacing it with our more lightweight TDTB module, which is inserted only after the fourth layer of the backbone network and more effectively utilizes temporal information, the model's performance is significantly improved.

\renewcommand{\arraystretch}{1.2}
\begin{table*}[!t]
\centering
\resizebox{\textwidth}{!}{
\begin{tabular}{c c c c c c c c c}
\toprule
 & \multicolumn{4}{c}{Components} & \multicolumn{4}{c}{LasHeR}\\
\cmidrule(lr){2-5}\cmidrule(lr){6-9}
 & TBSI & Dual-Template & TMCE & TDTB & SR & PR & NPR & FPS (2080Ti) \\
\midrule
\#1 & - & - & - & - & 53.2 & 65.9 & 62.6 & 47.04 \\
\#2 & \checkmark & - & - & - & 55.6 & 69.2 & 65.8 & 33.03 \\
\#3 & \checkmark & \checkmark & - & - & 55.9 & 69.8 & 66.1 & 29.71 \\
\#4 & - & \checkmark & - & \checkmark & 57.4 & 71.5 & 67.8 & 36.82 \\
\#5 & - & \checkmark & \checkmark & - & 56.1 & 70.4 & 66.6 & \textbf{47.48} \\
\#6 & \checkmark & \checkmark & \checkmark & - & 57.0 & 71.0 & 67.3 & 32.22 \\
\#7 & - & \checkmark & \checkmark & \checkmark & \textbf{58.0} & \textbf{72.3} & \textbf{68.5} & 40.65 \\
\bottomrule
\end{tabular}
}
\caption{Effectiveness comparison of the proposed BTMTrack components and the TBSI~\citep{hui2023bridging} module on the LasHeR dataset. Speed tests are conducted on a single RTX 2080Ti GPU. The best results are highlighted in \textbf{bold}.}
\label{tab:tab2}
\end{table*}

We denote \textbf{\#5} as the backbone network that applies the TMCE strategy with the dual-template design, but without the fusion module. \textbf{\#6} represents the case where the TBSI~\citep{hui2023bridging} module is added to \textbf{\#5}, while \textbf{\#7} represents our complete BTMTrack, where the TDTB module is incorporated into \textbf{\#5}. As shown by the metrics from \textbf{\#5} to \textbf{\#7}, introducing the TMCE strategy into the dual-template backbone significantly improves model speed while also enhancing performance to some extent. Adding the TBSI~\citep{hui2023bridging} module further improves performance. However, the TDTB module, which is more synergistic with TMCE, achieves the best results. Compared to the baseline network, our complete BTMTrack improves the Success Rate (SR) by 4.8\% and the Precision Rate (PR) by 6.4\%. Additionally, the introduction of the TMCE strategy reduces computational cost, compensating for the overhead introduced by the dual-template, thus striking an excellent balance between speed and performance.

\textbf{Comparison of Elimination Strategies}.
To evaluate the effectiveness of our proposed TMCE strategy, we conducted experiments on the LasHeR~\citep{li2021lasher} dataset under the same conditions, testing the changes in model performance when using different Elimination Strategies. \textit{\textbf{w/o any Strategy}} refers to removing the TMCE strategy entirely from BTMTrack, while \textit{\textbf{with CE}} replaces the TMCE strategy with the traditional CE~\citep{ye2022joint} strategy. Furthermore, to verify the contribution allocation discussed in Sec.\ref{sec:tmce}, which incorporates both multi-modal and temporal information into the Elimination Strategy, we designed two variants of the CE strategy as baselines for the ablation study: \textit{\textbf{with Add-CE}} directly adds the correlation maps between the templates and the search region from the two modalities and uses the resulting score to filter tokens; \textit{\textbf{with Max-CE}} directly takes the maximum value of the correlation maps between the templates and the search region from the two modalities and uses the resulting score for token filtering. \textit{\textbf{with TMCE}} represents the full BTMTrack model equipped with the TMCE strategy.

As shown in Tab.\ref{tab:tab3}, although the CE~\citep{ye2022joint} strategy does not improve performance, it slightly enhances the model’s speed. On the other hand, while the Add-CE and Max-CE strategies are faster, their incomplete filtering mechanisms introduce more background noise, leading to degraded performance. These two variants fail to fully utilize the dual-template information within the backbone network, which results in reduced model effectiveness. In contrast, the TMCE strategy significantly improves model performance while also enhancing speed to some extent, further demonstrating the superiority of the TMCE method.
\renewcommand{\arraystretch}{1.2}
\begin{table}[!t]
\centering
\resizebox{0.7\linewidth}{!}{
\begin{tabular}{l c c c c}
\toprule
Elimination Strategy & SR & PR & NPR & FPS (2080Ti) \\
\midrule
w/o \:any Strategy & 57.4 & 71.5 & 67.8 & 36.82 \\
with CE & 57.3 & 71.5 & 67.9 & 40.58 \\
with Add-CE & 56.7 & 70.6 & 67.0 & 41.65 \\
with Max-CE & 56.7 & 70.7 & 67.1 & \textbf{42.58} \\
with TMCE & \textbf{58.0} & \textbf{72.3} & \textbf{68.5} & 40.65 \\
\bottomrule
\end{tabular}
}
\caption{Performance comparison of the proposed TMCE with other elimination strategies on the LasHeR dataset.}
\label{tab:tab3}
\end{table}

\renewcommand{\arraystretch}{1.2}
\begin{table}[!t]
\centering
\resizebox{0.55\linewidth}{!}{
\begin{tabular}{ccc | ccc}
\toprule
\multicolumn{3}{c|}{\fontsize{8pt}{12pt}\selectfont{\textbf{Layers}}} & \multirow{2}{*}{\raisebox{-0.0ex}{\fontsize{8pt}{12pt}\selectfont \textbf{Success}}} & \multirow{2}{*}{\raisebox{-0.0ex}{\fontsize{8pt}{12pt}\selectfont \textbf{Precision}}}& \multirow{2}{*}{\raisebox{-0.0ex}{\fontsize{8pt}{12pt}\selectfont \textbf{NormPrec}}} \\
4 & 7 & 10  \\
\midrule
- & - & - & 56.1 & 70.4 & 66.6 \\
\checkmark & - & - & \textbf{58.0} & \textbf{72.3} & \textbf{68.5} \\
\checkmark & \checkmark & - & 56.2 & 70.2 & 66.5 \\
\checkmark & - & \checkmark & 57.0 & 71.5 & 67.4 \\
- & \checkmark & \checkmark & 54.9 & 68.9 & 65.0 \\
\checkmark & \checkmark & \checkmark & 57.8 & 72.1 & 68.3 \\
\bottomrule
\end{tabular}
}
\caption{Ablation study on the insertion layers of the proposed TDTB module on the LasHeR dataset.}
\label{tab:tab4}
\end{table}

\textbf{Inserting Layers of TDTB module}.
To explore the optimal insertion layer for the TDTB module within the model, we evaluated its effectiveness when inserted after different layers of the backbone network using the LasHeR dataset. The results, shown in Tab.\ref{tab:tab4}, indicate that inserting the TDTB module only after the 4th layer, compared to inserting it after the 4th, 7th, and 10th layers, not only maintains the model's performance but even slightly improves it. Furthermore, using a single insertion significantly reduces the number of parameters and improves the model's speed.

We analyzed the reasons behind this phenomenon: The TDTB module integrates four templates (two modalities and two temporal states) and performs six cross-modal attention operations. When inserted after the 4th layer, it better facilitates the integration of information from different modalities, enhancing the representation of mid-level features and improving contextual consistency. These mid-level features are already sufficient for the task. However, when TDTB is inserted after higher layers, such as the 7th or 10th layers, the model has already captured highly complex semantic information. Introducing a complex fusion module at this stage may disrupt the learned patterns, ultimately leading to a performance drop. Additionally, due to the TMCE strategy employed in BTMTrack, performing cross-modal fusion after token filtering in the earlier and middle stages of the network allows the model to fully utilize the most relevant tokens. At later stages, as the number of tokens decreases significantly, the remaining information is already well-refined and focused. Introducing complex cross-modal fusion operations at this point could lead to redundancy or conflicts in the information. Specifically, cross-modal fusion may involve information from different sources, and if this information is imprecise or conflicting, it could reduce the model's focus on core features.

Therefore, we adopt the design of inserting the TDTB module only after the 4th layer, which achieves optimal performance while reducing the model’s parameter count and striking an excellent balance between speed and performance.

\renewcommand{\arraystretch}{1.2}
\begin{table}[!t]
\centering
\resizebox{0.75\linewidth}{!}{
\begin{tabular}{l c c c c c}
\toprule
 & VIPT & BAT & TBSI & GMMT & BTMTrack \\
\midrule
NO  & 84.1/68.4 & 90.2/73.3 & \textbf{91.4}/74.1 & 90.9/\textbf{74.3} & 90.3/73.6 \\
PO  & 62.5/50.4 & 67.5/54.0 & 67.8/54.0 & 68.0/54.4 & \textbf{70.1}/\textbf{55.6} \\
TO  & 57.7/46.2 & 64.1/\textbf{51.1} & \textbf{64.3}/51.0 & 64.1/50.9 & \textbf{64.3}/50.7 \\
HO  & 46.8/43.4 & 56.5/51.0 & 60.6/\textbf{53.4} & 57.6/51.2 & \textbf{61.0}/\textbf{53.4} \\
MB  & 57.6/46.0 & 62.4/49.6 & 63.1/49.5 & 64.0/50.5 & \textbf{64.8}/\textbf{50.6} \\
LI  & 49.9/41.2 & 60.4/48.2 & 61.3/49.3 & 61.9/49.5 & \textbf{62.7}/\textbf{49.9} \\
HI  & 67.8/54.2 & 75.3/59.6 & 73.8/58.2 & 75.2/59.3 & \textbf{79.1}/\textbf{62.2} \\
AIV & 36.3/34.2 & 51.4/45.3 & \textbf{58.2}/\textbf{49.8} & 53.0/46.4 & 55.6/48.6 \\
LR  & 56.8/41.8 & 62.7/46.2 & 63.9/\textbf{47.3} & 63.7/\textbf{47.3} & \textbf{65.0}/\textbf{47.3} \\
DEF & 67.6/55.8 & \textbf{72.2}/58.5 & 71.6/58.7 & 70.8/58.0 & 72.1/\textbf{58.8} \\
BC  & 65.0/51.9 & 67.7/53.9 & \textbf{69.9}/\textbf{55.7} & 68.8/54.7 & 69.8/55.2 \\
SA  & 57.4/46.5 & 61.3/49.2 & 62.2/50.2 & 61.9/50.2 & \textbf{64.6}/\textbf{51.5} \\
CM  & 62.0/50.0 & 68.0/54.4 & 69.5/55.0 & 69.0/54.7 & \textbf{71.1}/\textbf{56.4} \\
TC  & 57.4/46.0 & 62.7/50.1 & 62.6/50.1 & 63.1/50.5 & \textbf{64.1}/\textbf{50.8} \\
FL  & 59.7/46.9 & 62.0/49.0 & 60.9/47.5 & \textbf{63.6}/\textbf{49.8} & 62.4/49.3 \\
OV  & 76.2/65.0 & \textbf{76.8}/\textbf{66.0} & 64.6/55.9 & 67.5/58.8 & 72.3/62.1 \\
FM  & 63.2/51.5 & 68.5/55.1 & 69.4/55.7 & 69.2/55.6 & \textbf{71.4}/\textbf{57.1} \\
SV  & 65.0/52.5 & 69.7/55.9 & 70.2/56.2 & 70.6/56.7 & \textbf{73.0}/\textbf{58.1} \\
ARC & 59.4/49.5 & 63.1/51.9 & 64.3/52.5 & 64.7/52.8 & \textbf{67.1}/\textbf{54.5} \\
\bottomrule
\end{tabular}
}
\caption{Attribute-based Precision/Success scores on the LasHeR dataset. All metrics are evaluated based on the original results published by the authors of the compared methods.}
\label{tab:tab5}
\end{table}

\subsection{Analysis and Visualization}
\textbf{Attribute-Based Performance}. To evaluate the performance of RGB-T tracking models under diverse challenges, the LasHeR dataset defines 19 attributes that represent real-world tracking difficulties. These attributes include no occlusion (NO), partial occlusion (PO), total occlusion (TO), hyaline occlusion (HO), motion blur (MB), low illumination (LI), high illumination (HI), abrupt illumination variation (AIV), low resolution (LR), deformation (DEF), background clutter (BC), similar appearance (SA), camera moving (CM), thermal crossover (TC), frame lost (FL), out-of-view (OV), fast motion (FM), scale variation (SV), and aspect ratio change (ARC). We compared our model with four of the most advanced models in recent years, including VIPT, BAT, TBSI, and GMMT, based on these attributes. The results, as shown in Tab.\ref{tab:tab5}, demonstrate that our model achieves the best performance across most scenarios. Notably, it exhibits significant advantages in conditions such as fast motion, scale variation, aspect ratio change, thermal crossover, high illumination, and low illumination. This further highlights the importance of integrating temporal information and cross-modal information for feature fusion and selection.

\begin{figure*}[!htb]
\centering 
\includegraphics[width=1\textwidth]{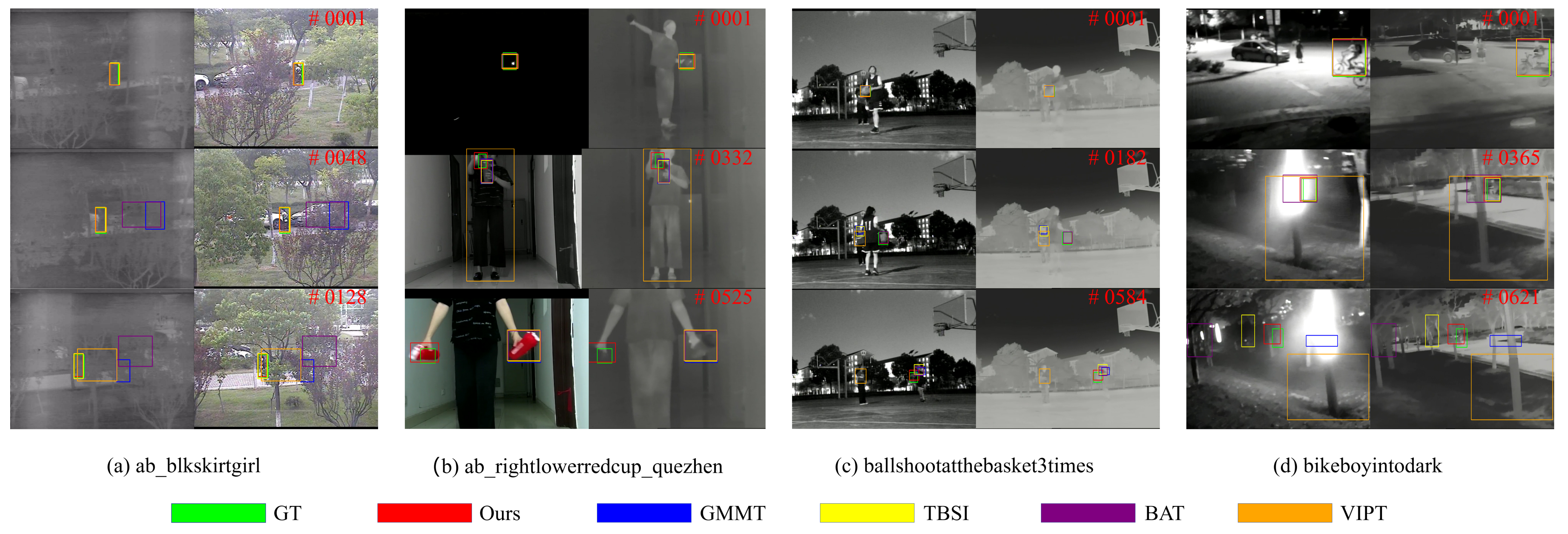}
\caption{Qualitative comparison of our method with other RGB-T trackers on four representative sequences from the LasHeR dataset.} \label{sample-figure}
\label{fig:fig4}
\end{figure*}
\textbf{Visualization}. As shown in Fig.\ref{fig:fig4}, we visually compared our model with four of the most advanced RGB-T trackers in recent years. To better demonstrate the robustness of our model, we selected four representative video sequences from the LasHeR dataset, which include challenges such as occlusion, low illumination, high illumination, and fast motion. For example, in the first video sequence, a girl walking on the road is partially occluded by trees; the second sequence presents the challenge of a low-light environment; in the third sequence, a basketball moves at high speed; and in the fourth sequence, high illumination creates significant difficulty in determining the position of a boy riding a bicycle. The results show that our model performs exceptionally well under a variety of challenging conditions.

\section{Conclusion}
In this paper, we propose an efficient cross-modal interaction method for RGB-T tracking that integrates dynamic templates into the backbone network. Our approach aims to fully exploit temporal and cross-modal information. Unlike previous methods, which either overlook or fail to effectively utilize temporal information, BTMTrack leverages the TMCE strategy to focus the model on target-related features, reducing computational overhead while avoiding irrelevant background noise. Additionally, we introduce the TDTB module to further enhance the interaction between templates and the search region by extracting more robust target-related contextual information. Evaluations on three widely used RGB-T benchmark datasets demonstrate that our method accurately locates targets and achieves state-of-the-art performance.

\section*{Acknowledgments}
This work is supported in part by National Research and Development Key Project No. 2023YFF0905603, and in part by National Joint Fund Key Project (NSFC- Guangdong Joint Fund) under Grant U21A20478.

\bibliographystyle{main} 
\bibliography{main}

\end{document}